%% file: main.tex
\documentclass[letterpaper]{article}

\usepackage{natbib,alifeconf}  %
\usepackage{url,hyperref}
\usepackage{amsmath,amssymb}
\usepackage{graphicx}
\usepackage{subcaption}
\usepackage{booktabs}
\usepackage{xcolor}

\title{Diffusing Blame: Task-Dependent Credit Assignment \\in Biologically Plausible Dual-Stream Networks}

\author{
\input{paper_authors.tex}
}

\begin{document}
\maketitle

\begin{abstract}
Biological neural circuits obey Dale's principle: each neuron's synapses are uniformly excitatory or inhibitory. Artificial networks that respect this constraint must coordinate separate excitatory and inhibitory populations, fundamentally changing how credit is assigned during learning. Several biologically plausible learning rules avoid backpropagation's weight transport requirement, but it has been difficult to achieve strong performance under Dale's principle beyond MNIST. 
Error Diffusion (ED) was originally proposed in a dual-stream excitatory/inhibitory architecture, where learning is driven by routing global error signals to all layers without transporting transposed forward weights or relying on random feedback matrices.
Whether such a rule can scale under Dale's principle across both supervised classification and reinforcement learning remains unknown. 
Here, we introduce modulo error routing to extend Error Diffusion beyond binary classification, and show that a dual-stream excitatory/inhibitory architecture trained with this method achieves $96.7\%$ on MNIST and establishes a $61.7\%$ baseline on CIFAR-10, demonstrating that representation learning is possible even when strictly enforcing Dale’s principle.
For the classification setting, we introduce three domain-specific innovations: layer-specific sigmoid widths, batch-centered class error signals, and asymmetric initialization, and ablation analysis reveals that their relative importance \emph{reverses} between MNIST and CIFAR-10, exposing task-dependent credit-assignment bottlenecks invisible to single-benchmark evaluation. 
In reinforcement learning, we integrate ED with Proximal Policy Optimization (PPO) and evaluate it on continuous-control tasks in Google Brax and on Craftax, an open-ended exploration task. We show that ED-PPO achieves competitive performance relative to Direct Feedback Alignment, a backpropagation-free baseline.
\end{abstract}

\section{Introduction}

In biological neural circuits, the separation of excitatory and inhibitory neuronal populations is a fundamental organizational principle, formalized as Dale's principle: each neuron releases the same neurotransmitter at all of its synapses \citep{dale1935pharmacology, eccles1976dale, strata1999dale}. This constraint shapes cortical computation through balanced excitation and inhibition
\citep{markram2004interneurons}. 
Artificial neural networks, by contrast, freely assign positive and negative weights to any connection, a simplification that enables efficient credit assignment via backpropagation but bears little resemblance to biological learning.

Training deep networks with backpropagation requires the backward pass to use exact transposes of forward weight matrices, the \emph{weight transport problem}, which lacks biological support \citep{crick1989recent, bengio2015towards}. Several alternatives address this concern. Feedback Alignment (FA) replaces backward weights with fixed random matrices \citep{lillicrap2016random}, Direct Feedback Alignment (DFA) routes output errors directly to each hidden layer through fixed random matrices \citep{nokland2016direct}, 
and predictive coding frameworks approximate backpropagation through local Hebbian updates \citep{whittington2017approximation, sacramento2018dendritic}. While these methods avoid weight transport, none enforce Dale's principle: all allow arbitrary sign weights. Error Diffusion (ED), originally proposed by \citep{kaneko2000error_diffusion}, is a local learning rule in which weight updates depend on presynaptic activity, a postsynaptic activation derivative, and a single global error sign \citep{fujita2026diagnostic}. This locality makes ED naturally compatible with biological constraints, but prior work demonstrated its effectiveness only on binary classification and a simple classification task like MNIST \citep{fujita2026diagnostic}.

The central question we address is whether a biologically plausible architecture that enforces Dale's principle can achieve competitive performance across both supervised classification and reinforcement learning, and what the task-dependent importance of its components reveals about credit assignment under biological constraints. We make three contributions:
\begin{enumerate}
\item A dual-stream excitatory/inhibitory ED architecture that maintains non-negative weights across classification (MNIST, CIFAR-10) and reinforcement learning (Brax locomotion tasks, Craftax) by extending the original ED via modulo error routing.
\item For the classification setting, three domain-specific innovations: layer-specific sigmoid widths, batch-centered class error, and asymmetric initialization, whose cross-task ablation reveals that credit-assignment bottlenecks shift qualitatively with task difficulty.
\item A cross-domain evaluation showing that ED-PPO achieves returns comparable to backpropagation-based and DFA-based PPO on continuous control tasks such as Ant, Humanoid, and HalfCheetah. On the more complex Craftax task, ED-PPO achieves performance comparable to DFA.
\end{enumerate}

\begin{figure*}[t]
    \centering
    \includegraphics[width=0.9\textwidth]{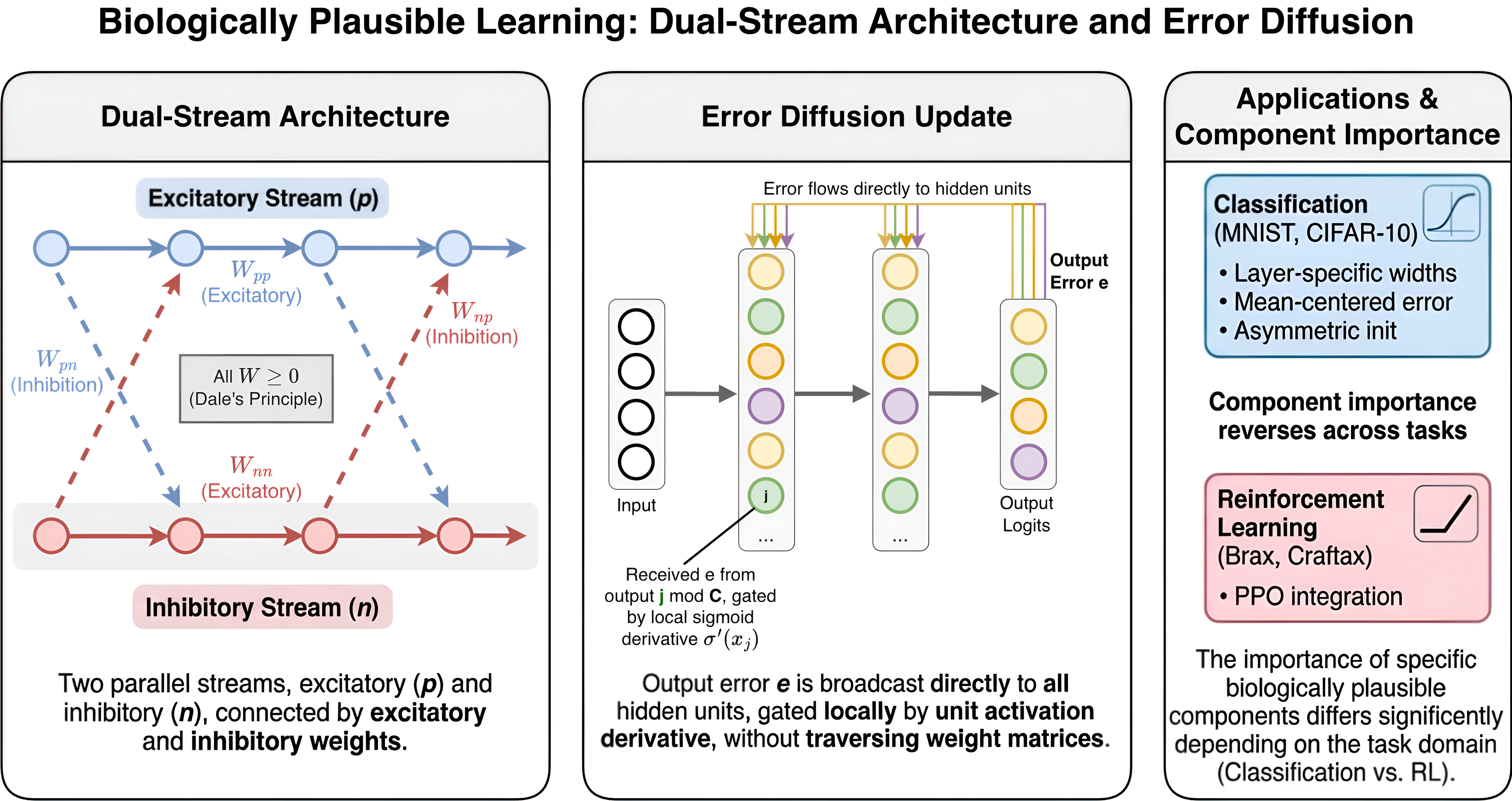}
    \caption{Overview of the dual-stream Error Diffusion framework. \textbf{Left:} The excitatory/inhibitory architecture maintains separate streams ($\mathbf{p}$, $\mathbf{n}$) with four non-negative weight matrices per layer, enforcing Dale's principle structurally. \textbf{Center:} The Error Diffusion update broadcasts output error directly to hidden layers without weight transport or random feedback matrices. \textbf{Right:} The shared architecture is applied to classification (with domain-specific sigmoid widths, batch-centered class error, and asymmetric initialization) and reinforcement learning (with PPO integration).}
    \label{fig:concept}
\end{figure*}

\section{Related Work}

\paragraph{Biologically plausible learning rules.}
The weight transport problem has motivated a family of alternatives to backpropagation \citep{crick1989recent, bengio2015towards}. Feedback Alignment uses fixed random backward weights \citep{lillicrap2016random}, and \citet{liao2016important} showed that approximate weight symmetry can support learning in deeper networks. DFA projects output errors directly to hidden layers \citep{nokland2016direct}, and \citet{launay2020direct} demonstrated that DFA scales to modern architectures, including transformers and convnets. Difference target propagation computes layer-wise learning targets instead of gradients \citep{lee2015difference}. Predictive coding provides a complementary framework through local Hebbian updates \citep{whittington2017approximation}, and dendritic cortical microcircuit models approximate backpropagation with segregated dendrites \citep{sacramento2018dendritic}. \citet{bartunov2018assessing} found that biologically motivated algorithms generally struggle to scale beyond simple benchmarks, and \citet{richards2019deep} argued that biological plausibility and computational performance need not be in tension if the right architectural principles are identified. \citet{filipovich2022silicon} demonstrated DFA on photonic hardware, showing that bioplausible rules can enable novel computing substrates.

\paragraph{Dale's principle in neural networks.}
\citet{song2016training} introduced a framework for training excitatory-inhibitory recurrent networks with Dale's principle, separating synaptic magnitudes from signs by parameterizing weights as
$W = W^{+} \odot M$, where $W^{+} \geq 0$ contains trainable connection magnitudes and $M$ is a fixed sign mask encoding each unit's excitatory or inhibitory identity.
\citet{cornford2021learning} proposed Dale's ANNs, feedforward architectures with separate excitatory and inhibitory populations that can learn comparably to standard ANNs by treating inhibition as a normalization mechanism.
\citet{grossberg1987competitive} explored competitive learning in networks with separate excitatory and inhibitory interactions.
Error Diffusion \citep{fujita2026diagnostic, kaneko2000error_diffusion} is also naturally Dalean, representing positive and negative components through separate non-negative populations while avoiding both weight transport and random feedback.
Building on this framework, our work extends Error Diffusion to multi-class classification and reinforcement learning.

\paragraph{Reinforcement learning without backpropagation.}
Evolution Strategies (ES) provide a gradient-free alternative to policy gradient methods \citep{salimans2017evolution}, but scale poorly with parameter count. Proximal Policy Optimization (PPO) is a widely used policy gradient method \citep{schulman2017proximal} that typically relies on backpropagation for gradient computation. Biologically plausible alternatives for RL credit assignment have been proposed---reward-modulated Hebbian rules can solve decision-making tasks by gating synaptic updates with a global reward signal \citep{pfeiffer2010reward}, and \citet{izhikevich2007solving} showed that linking spike-timing-dependent plasticity with dopamine signaling addresses the distal reward problem---but these methods have not been demonstrated with deep networks on continuous-control benchmarks. We replace backpropagation in PPO's gradient computation with Error Diffusion, evaluating on Brax physics tasks \citep{freeman2021brax} and Craftax \citep{matthews2024craftax}.

\section{Method}

\subsection{Dual-Stream Architecture}

We adopt the original Error Diffusion formulation proposed by \citep{kaneko2000error_diffusion}.
To enforce Dale's principle, 
we split each layer into an excitatory (\textbf{p}) and inhibitory (\textbf{n}) stream.  
The forward pass for layer $i$ computes:
\begin{align*}
\mathbf{p}_i &= \phi_i\!\left(+\mathbf{p}_{i-1} W_{pp} - \mathbf{n}_{i-1} W_{np} + \mathbf{b}_p\right) \\
\mathbf{n}_i &= \phi_i\!\left(+\mathbf{n}_{i-1} W_{nn} - \mathbf{p}_{i-1} W_{pn} + \mathbf{b}_n\right)
\end{align*}
where all weight matrices $W_{pp}, W_{np}, W_{nn}, W_{pn} \geq 0$ element-wise, bias parameters $\mathbf{b}_p$ and $\mathbf{b}_n$ are not necessary non-negative, $\phi_i$ is a layer-specific activation function. The negation signs in front of $W_{np}$ and $W_{pn}$ are structural (hardcoded), ensuring that cross-stream connections are inhibitory while all learnable parameters remain non-negative. This dual-stream design requires four weight sub-matrices per layer, resulting in ${\sim}4\times$ more parameters than an unconstrained single-stream network of the same width (e.g., ${\sim}$32M vs ${\sim}$8M for DFA on the same architecture).

\subsection{Error Diffusion Update Rule}

\begin{figure*}[t]
    \centering
    \includegraphics[width=0.9\textwidth]{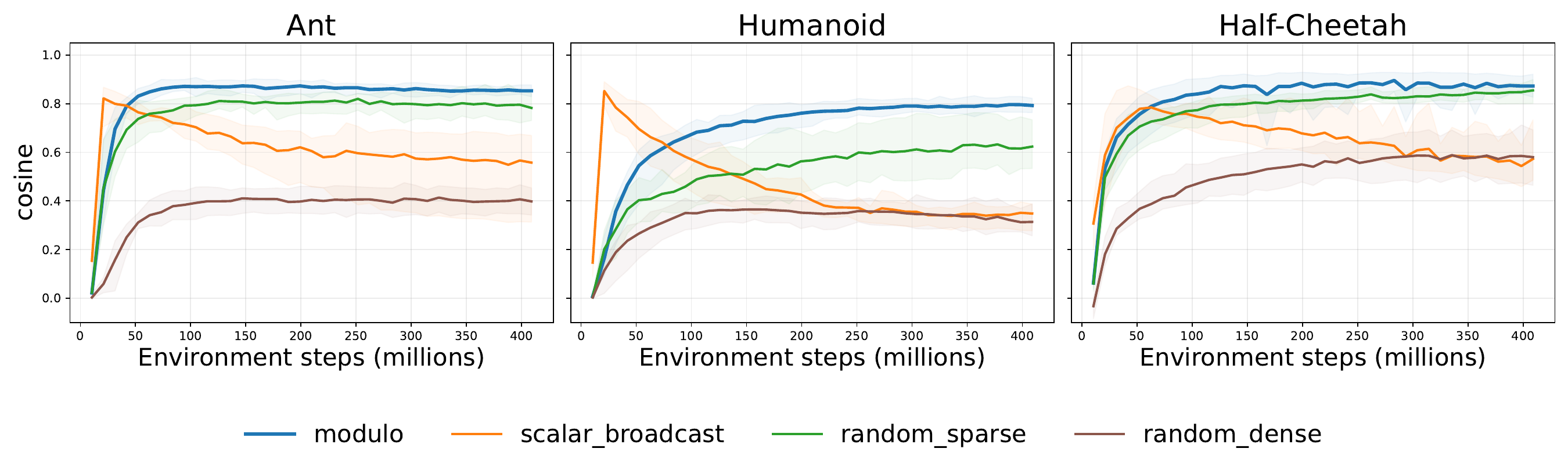}
    \caption{Cosine similarity between the ED local update direction and the BP gradient direction, restricted to policy trunk parameters and excluding the policy in a dual-stream architecture, over PPO training on Ant, Humanoid, and Half-Cheetah. Lines show the mean over 5 seeds; shaded regions show the seed-wise min-max range. \texttt{modulo} uses deterministic one-output-per-hidden-unit routing, \texttt{random\_sparse} uses random one-output-per-hidden-unit routing, \texttt{scalar\_broadcast} sends the same normalized sum of output errors to every hidden unit, and \texttt{random\_dense} denotes the normalized dense random feedback variant used in \citep{launay2019principledtrainingneuralnetworks}. Higher cosine indicates closer agreement between ED and BP trunk parameter-gradient directions.}
    \label{fig:cosine_similarity}
\end{figure*}

The ED update replaces backpropagated layerwise errors with an output-space error signal that is routed directly to hidden units. While the original formulation was developed for binary classification \citep{kaneko2000error_diffusion, fujita2026diagnostic}, we extend ED to multi-output prediction by assigning each hidden unit a fixed output channel. For a hidden unit $i$, we define
$r(i) = i \bmod C$, where $C$ is the output dimension, and use the routed error component $s_{r(i)}$ as its learning signal. (For other design choices of this error routing and why, see Figure \ref{fig:cosine_similarity}) 

For a mini-batch, let $S \in \mathbb{R}^{B \times C}$ denote the output error signal, and let
$M \in \{0,1\}^{H \times C}$ be the fixed routing matrix with
$M_{ic}=1$ iff $r(i)=c$. 
The routed hidden error is
\begin{equation}
R = S M^\top,
\end{equation}
and the corresponding matrix of local postsynaptic drives is
\begin{equation}
U_p = \phi'(Z_p) \odot R,
\end{equation}
where $Z_p \in \mathbb{R}^{B \times H}$ is the positive-stream preactivation matrix.

Let $A_p \in \mathbb{R}^{B \times K}$ denote the positive-stream presynaptic activations feeding the layer. In fully connected notation, the positive-to-positive ED update is
\begin{equation}
\Delta W_{pp}
\propto
A_p^\top U_p.
\label{eq:update}
\end{equation}

In the supervised image classifiers, $C=10$ and $S$ is the batch-centered one-vs-all classification error. In Craftax policy networks, $C$ is the number of discrete action logits. In MuJoCo policy networks, $C$ is the policy distribution parameter dimension for the TanhNormal policy.

Each local weight update is proportional to presynaptic activity, and this routed postsynaptic drive, with stream-specific signs for the dual positive/negative pathways. 
This deterministic modulo routing provides coarse output-associated credit assignment without transporting forward weights or using random feedback matrices. 
Unlike DFA, whose fixed feedback matrices are random, ED uses a structured correspondence between hidden units and output dimensions.

\subsection{Classification-Specific Innovations}

For the classification setting (MNIST, CIFAR-10), we introduce three domain-specific innovations that address failure modes specific to multi-class classification under Dale's principle. These three innovations are specific to the classification setting and are not used in the RL extension.

\paragraph{Layer-specific sigmoid widths.} %
In the classification architecture, $\phi_i(z) = 1/(1 + e^{-2z/\alpha_i})$ with layer-type-specific sigmoid width $\alpha_i$.
The vanishing gradient problem in sigmoid networks is well documented \citep{roodschild2020new, ding2018activation}, and various parameterized activations have been proposed \citep{szandala2020review, hammad2024deep}. Since the sigmoid derivative directly gates the error signal in ED (Eq.~\ref{eq:update}), gradient attenuation is especially severe: post-hoc analysis reveals a $25\times$ decay from output to first hidden layer. Wider sigmoids (larger $\alpha_i$) maintain larger derivatives, preventing premature saturation. 
We set $\alpha = 3.0$ for CIFAR-10 convolutional layers and $\alpha = 6.0$ for fully connected layers, including the output layer; MNIST uses only the $\alpha = 6.0$ fully connected setting.

\paragraph{Batch-centered class error signals.}
For the 10-way classification tasks, the ED update uses independent sigmoid output activations rather than a softmax. 
For a mini-batch of size $B$, we form a one-vs-all output error
$E_{b,c} = \mathbf{1}[y_b=c] - \hat{y}_{b,c}$, where $\hat{y}_{b,c}$ is the positive-stream output activation for class $c$. Since each example contributes one target class and nine non-target classes, the raw one-vs-all error can contain a strong class-wise offset, especially early in training. We therefore subtract the mini-batch mean separately for each class:
$\tilde{E}_{b,c} = E_{b,c} - \frac{1}{B}\sum_{b'=1}^{B} E_{b',c}$.
This makes the error signal zero-mean across the batch for every class, reducing persistent suppression or excitation of class channels caused by the one-vs-all target imbalance. 
The centered signal is then applied directly to the output layer, with the negative stream receiving the opposite sign; hidden fully connected units and convolutional channels receive class-routed copies of the same centered error.

\paragraph{Asymmetric E/I initialization.}
For convolutional and fully connected layers, weight parameters are initialized from a non-negative uniform distribution in $[0,1)$ and scaled by the inverse square root of the fan-in. 
Hidden excitatory weights ($W_{pp}, W_{nn}$) are then scaled by $1.5\times$, while inhibitory weights ($W_{np}, W_{pn}$) are scaled by $0.5\times$, giving an expected E/I scale ratio of $3:1$. 
The final fully connected output layer uses symmetric initialization ($1.0\times$ for both excitatory and inhibitory weights) to avoid output saturation.

\subsection{RL Extension: ED-PPO}

For reinforcement learning, we integrate a Dale-constrained dual-stream Error Diffusion (ED) architecture into PPO. 
The policy and value networks are both dual-stream MLPs with separate positive and negative pathways and non-negative synaptic weights. Each layer computes excitatory-minus-inhibitory preactivations for the positive and negative streams, applies a width-scaled sigmoid nonlinearity, and combines the final streams as $\hat{y} = y^+ - y^-$. Both Dale implementations initialize the two streams symmetrically from the same observation, $\mathbf{p}_0 = \mathbf{n}_0 = x$, after optional stream normalization, rather than splitting inputs with ReLU.

ED replaces backpropagation through hidden layers after the PPO's objective function supplies an output-level error signal. For vector-valued policy outputs, the error is routed to hidden units by output class/channel assignment; for the scalar value network, the error is broadcast to all hidden units. All weight matrices are initialized non-negative using absolute Gaussian samples, with separate scaling for excitatory and inhibitory pathways, and weights are clamped non-negative after optimizer updates.

Figure \ref{fig:concept} summarizes the error diffusion training procedure and methodological contributions.

\subsection{Experimental Protocol}

\paragraph{Classification.} We evaluate six variants on MNIST and CIFAR-10: (1)~the \textbf{proposed ED} with all three innovations, (2)~a \textbf{DFA baseline} using fixed random feedback matrices without Dale's constraints, (3)~the \textbf{seed ED} (the dual-stream architecture without any of the three innovations, i.e., using uniform sigmoid width $\alpha=1.0$, raw error signals, and symmetric initialization), and three ablations removing one innovation each: (4)~\textbf{no batch-centered class error}, (5)~\textbf{symmetric init}, and (6)~\textbf{uniform sigmoid width}. Each variant is trained for 250 epochs with five random seeds per task (60 runs total). For classification, weights are clamped to a floor of $10^{-4}$ after each update to enforce Dale's principle.

\paragraph{Reinforcement learning.} We compare ED-PPO against BP-PPO (standard backpropagation), DFA-PPO (Direct Feedback Alignment)~\citep{nokland2016direct}, and ES~\citep{salimans2017evolution} on three Brax ~\citep{freeman2021brax} locomotion tasks (Ant, HalfCheetah, Humanoid). On Craftax~\citep{matthews2024craftax}, we compare against BP-PPO and DFA-PPO. 
Additionally, we evaluate a non-Dalean variant of ED-PPO, in which weights are not constrained to be non-negative, and denote it as ``ED-PPO (non-Dalean)''.
Each algorithm is evaluated with five random seeds per environment. We report the final episode reward (the reward at the last evaluation checkpoint) averaged over seeds; statistical comparisons use Welch's $t$-test with $\alpha = 0.05$. The RL experiments use a separate training and evaluation pipeline from the classification experiments, with both pipelines sharing the same dual-stream ED architecture but differing in activation functions, normalization, and optimization procedure as described above.

\section{Results}

\subsection{Classification: Main Comparison}

Figure~\ref{fig:ablation} summarizes accuracy across all six classification variants. The proposed ED achieves $96.7 \pm 0.1\%$ on MNIST and $61.7 \pm 0.7\%$ on CIFAR-10, a substantial improvement over seed ED ($50.4 \pm 9.8\%$ and $11.6 \pm 2.2\%$). DFA achieves higher accuracy on both tasks ($97.6\%$ and $69.1\%$) but violates Dale's principle, requiring ${\sim}2.84$M negative weights. The accuracy gap between ED and DFA widens from $0.9$ pp on MNIST to $7.4$ pp on CIFAR-10, suggesting that the cost of maintaining non-negative weights grows with task difficulty.
We note that ${\sim}62\%$ test accuracy on CIFAR-10 is still far from competitive compared to traditional gradient-based methods. That being said, this is the first time ED has been successfully applied to convolutional neural networks. Previously, \citet{fujita2026diagnostic} obtained ${\sim}55.2\%$ test accuracy on CIFAR-10 using a purely feedforward MLP with flattening.

\subsection{Classification: Cross-Task Ablation Reversal}

\begin{figure*}[t]
    \centering
    \includegraphics[width=\textwidth]{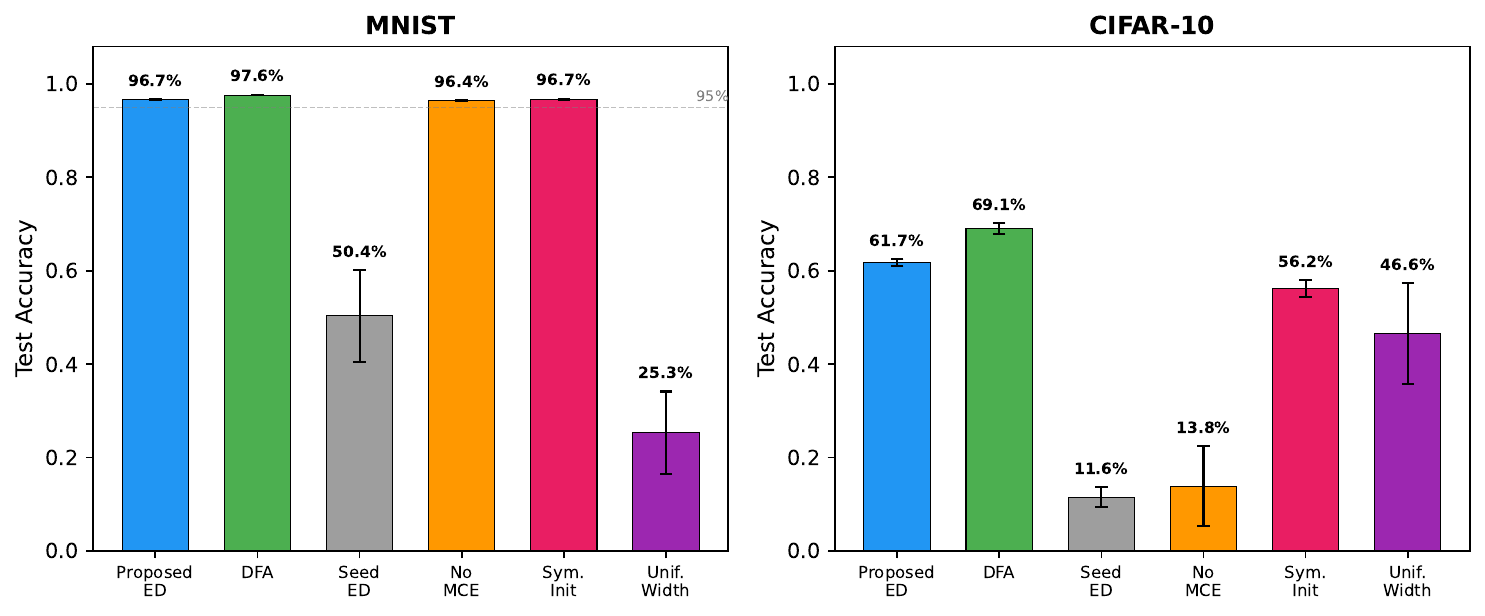}
    \caption{Classification accuracy across all six variants on MNIST (left) and CIFAR-10 (right). Error bars: $\pm 1$ std over 5 seeds. The ablation importance hierarchy reverses between tasks: layer-specific widths dominate on MNIST, batch-centered class error dominates on CIFAR-10.}
    \label{fig:ablation}
\end{figure*}

In the classification setting, ablation analysis reveals that the importance hierarchy of the three innovations reverses between tasks (Figure~\ref{fig:ablation}, Table~\ref{tab:deltas}).

\paragraph{MNIST.} Removing layer-specific sigmoid widths is catastrophic ($96.7\% \to 25.3\%$, $\Delta = -71.4$ pp), collapsing accuracy to near-chance. Removing batch-centered class error causes only $\Delta = -0.3$ pp, and symmetric initialization has no measurable effect ($\Delta = +0.0$ pp). Gradient flow regulation is the sole bottleneck on this easier task.

\paragraph{CIFAR-10.} The hierarchy reverses. Removing batch-centered error is now the most destructive ablation ($61.7\% \to 13.8\%$, $\Delta = -47.9$ pp), causing four of five seeds to collapse. Uniform sigmoid width causes $\Delta = -15.1$ pp, and symmetric initialization causes $\Delta = -5.5$ pp. All three innovations contribute, but their relative ordering changes fundamentally.

\begin{table}[t]
\centering
\caption{Ablation effect sizes ($\Delta$ = proposed ED minus ablation, in percentage). Negative values indicate accuracy loss.}
\label{tab:deltas}
\small
\begin{tabular}{@{}lrr@{}}
\toprule
Removed Component & MNIST $\Delta$ & CIFAR-10 $\Delta$ \\
\midrule
Batch-Centered Error & $-0.3$ & $-47.9$ \\
Layer-Specific Widths & $-71.4$ & $-15.1$ \\
Asymmetric Init & $+0.0$ & $-5.5$ \\
\bottomrule
\end{tabular}
\end{table}

\paragraph{Interpretation.} This reversal reflects qualitatively different credit-assignment bottlenecks. MNIST's well-separated features allow learning even without error centering; but without wide sigmoids, the derivatives vanish entirely. 
On CIFAR-10, higher inter-class similarity makes the 9:1 error imbalance overwhelming:
batch-centering becomes essential to prevent uniform output suppression. 
The reversal demonstrates that evaluating biologically plausible methods on a single benchmark may obscure critical design trade-offs.

\subsection{Reinforcement Learning: Cross-Domain Generalization}

\begin{figure*}[t]
    \centering
    \begin{subfigure}[t]{0.32\textwidth}
        \centering
        \includegraphics[width=\linewidth]{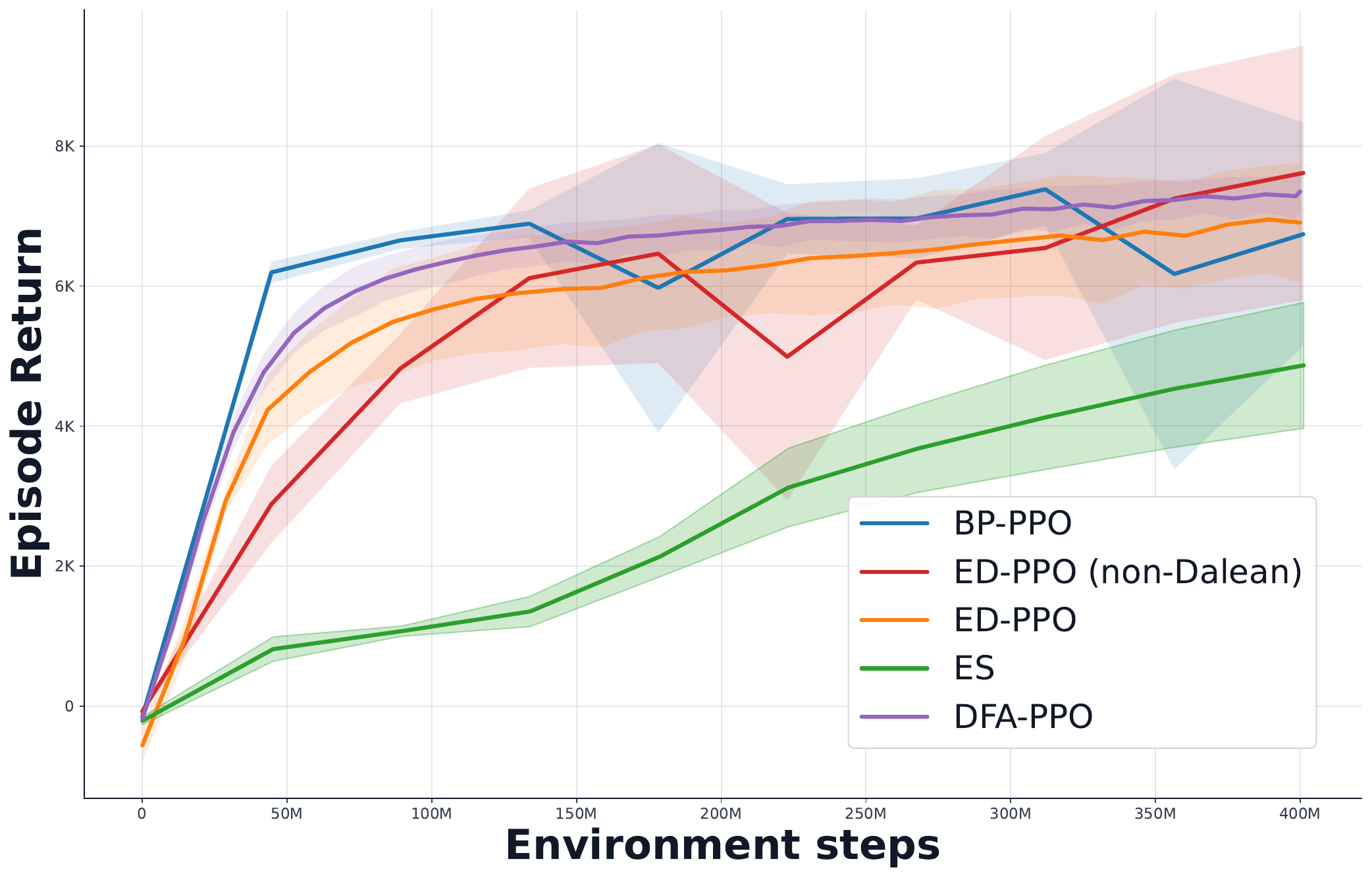}
        \caption{Ant}
        \label{fig:rl_ant}
    \end{subfigure}
    \hfill
    \begin{subfigure}[t]{0.32\textwidth}
        \centering
        \includegraphics[width=\linewidth]{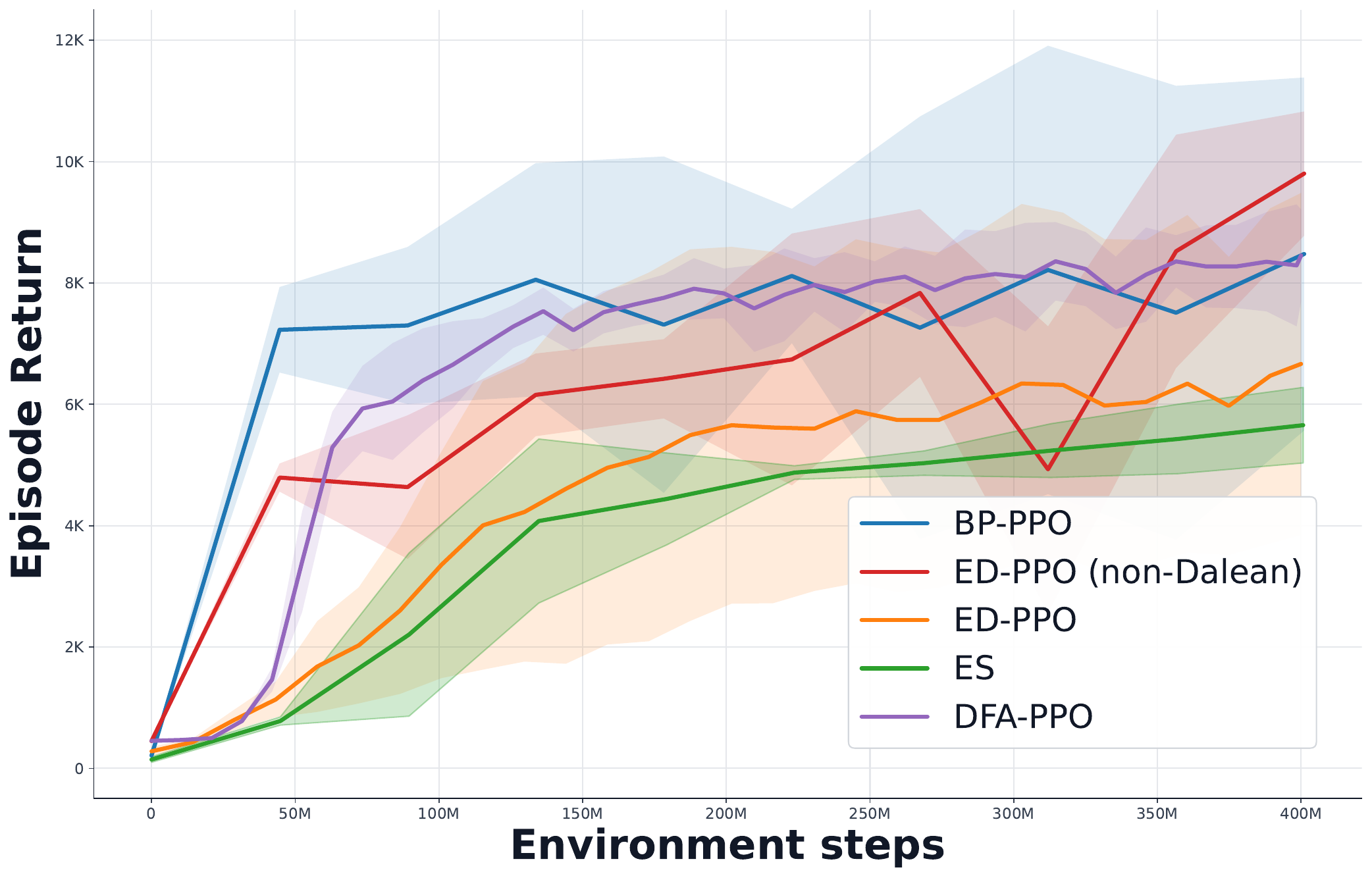}
        \caption{Humanoid}
        \label{fig:rl_humanoid}
    \end{subfigure}
    \hfill
    \begin{subfigure}[t]{0.32\textwidth}
        \centering
        \includegraphics[width=\linewidth]{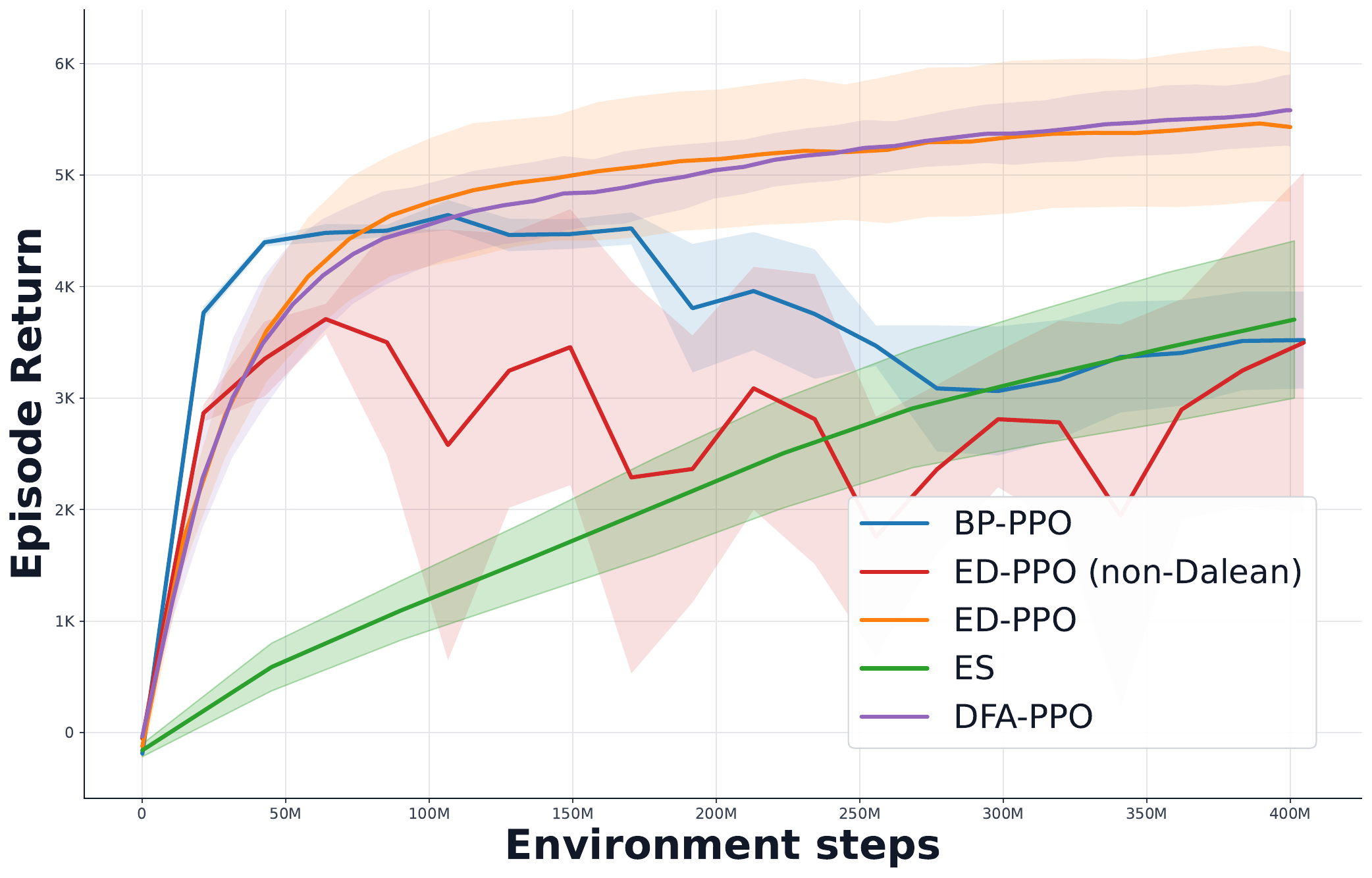}
        \caption{HalfCheetah}
        \label{fig:rl_halfcheetah}
    \end{subfigure}

    \caption{Episode return across three RL environments. Error bars indicate $\pm 1$ standard deviation over 5 seeds.} %
    \label{fig:rl}
\end{figure*}

\begin{figure}[t]
    \centering
    \includegraphics[width=\columnwidth]{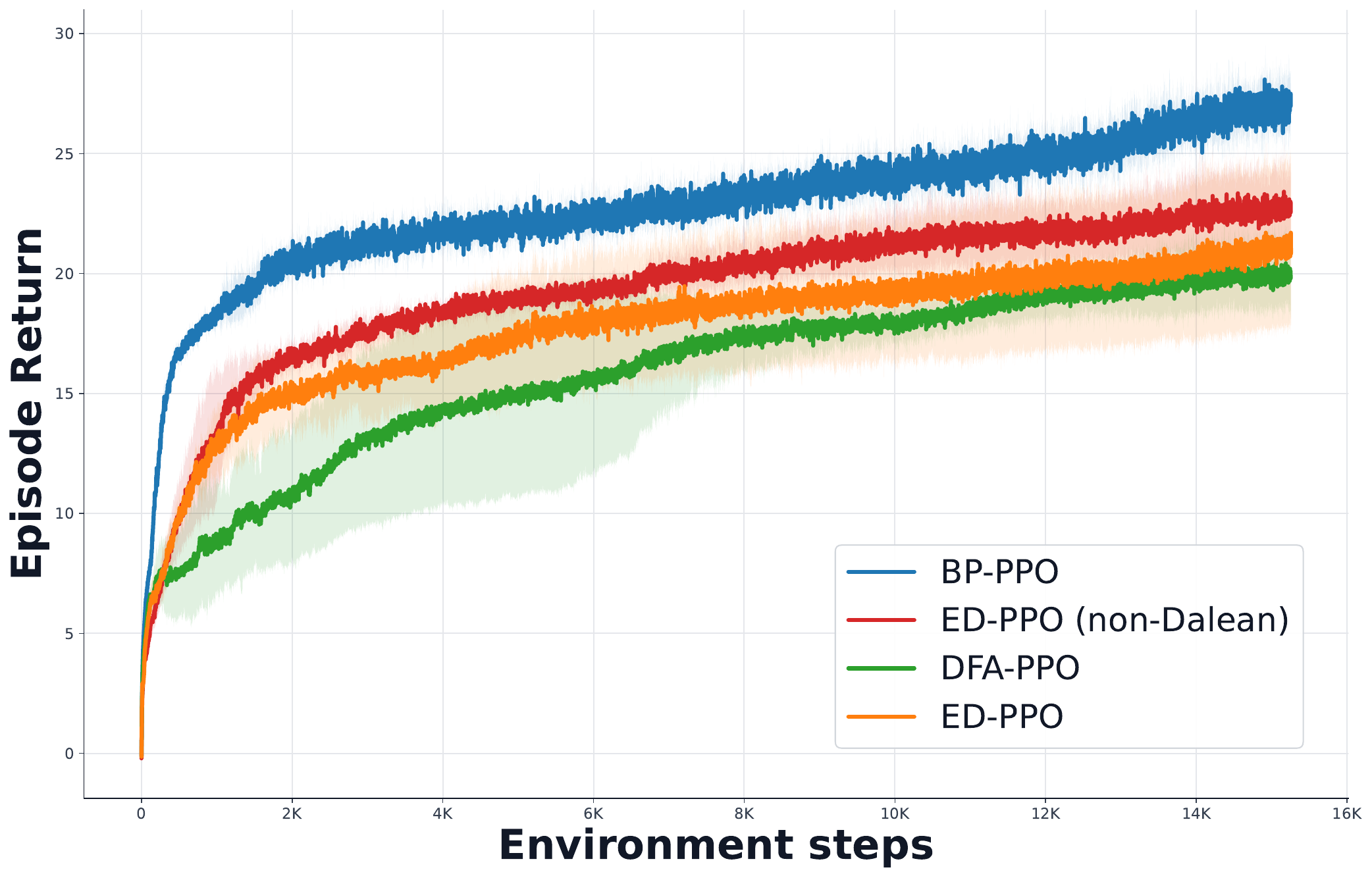}
    \caption{Episode return on Craftax comparing ED-PPO, BP-PPO, and DFA. Error bars: $\pm 1$ std over 5 seeds.}
    \label{fig:craftax}
\end{figure}

Figures~\ref{fig:rl} and~\ref{fig:craftax} report performance across four RL tasks. ED-PPO is strongest on HalfCheetah, where it outperforms BP-PPO ($5494 \pm 691$ vs $3520 \pm 485$; $p < 0.001$), ED-PPO (non-Dalean) ($p = 0.028$), and ES ($p = 0.003$), while matching DFA-PPO ($5581 \pm 359$). On Ant it is on par with both PPO variants ($6891 \pm 835$), significantly exceeding only ES ($p = 0.004$). On Humanoid ($6670 \pm 2592$) and Craftax ($20.9 \pm 2.9$) it trails BP-PPO and ED-PPO (non-Dalean), though the Craftax gap to the latter is not significant.

ED-PPO (non-Dalean) also performs competitively across the same four environments, on par with BP-PPO on Humanoid ($9804 \pm 1144$ vs $8478 \pm 3252$). On Ant, ED-PPO (non-Dalean) achieves a higher mean ($7616 \pm 2031$ vs $6740 \pm 1781$) but the difference is not significant due to high variance. On HalfCheetah, ED-PPO (non-Dalean) achieves a similar score ($3498 \pm 1703$ vs $3520 \pm 485$). Notably, DFA-PPO on Craftax achieves only $19.8 \pm 1.5$, significantly below ED-PPO (non-Dalean) ($p = 0.010$), demonstrating that random feedback pathways that suffice for simpler tasks fail on complex, open-ended environments.

ED-PPO (non-Dalean) exhibits higher variance than BP-PPO on Ant and HalfCheetah, though on Humanoid its variance is lower (1144 vs 3252), suggesting that the ED gradient signal can introduce additional stochasticity. 
The Craftax shortfall ($-4.0$, $p = 0.002$) and higher variance across environments suggest that the coarse modular error routing of ED may be less reliable for tasks requiring fine-grained temporal credit assignment or stable convergence.

\subsection{DFA Across Domains}

DFA provides an informative comparison point across both classification and RL. In classification, DFA achieves the highest accuracy (97.6\% MNIST, 69.1\% CIFAR-10) but violates Dale's principle. On Craftax, DFA-PPO is the weakest method ($19.8$ vs BP-PPO $27.0$, ED-PPO $23.0$), demonstrating that random feedback pathways that work for supervised learning can fail for complex RL. This cross-domain pattern---DFA strong on classification, weak on complex RL---parallels the finding that method components assume different importance under different task demands.

\subsection{Post-hoc Analysis}

\begin{figure*}[t]
    \centering
    \begin{subfigure}[t]{0.48\textwidth}
        \centering
        \includegraphics[width=\textwidth]{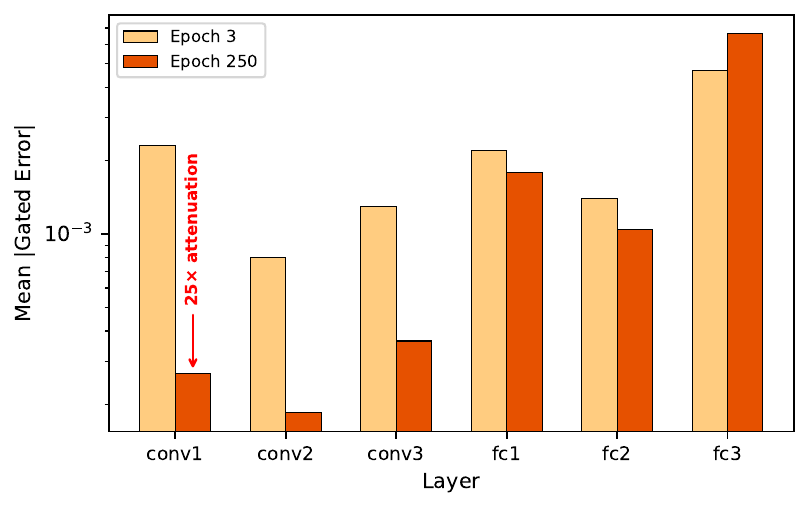}
        \caption{Per-layer local surrogate-gradient magnitudes at epochs 3 and 250 on CIFAR-10 (log scale). The $25\times$ attenuation from output to first hidden layer motivates layer-specific sigmoid widths.}
        \label{fig:gradient}
    \end{subfigure}
    \hfill
    \begin{subfigure}[t]{0.48\textwidth}
        \centering
        \includegraphics[width=\textwidth]{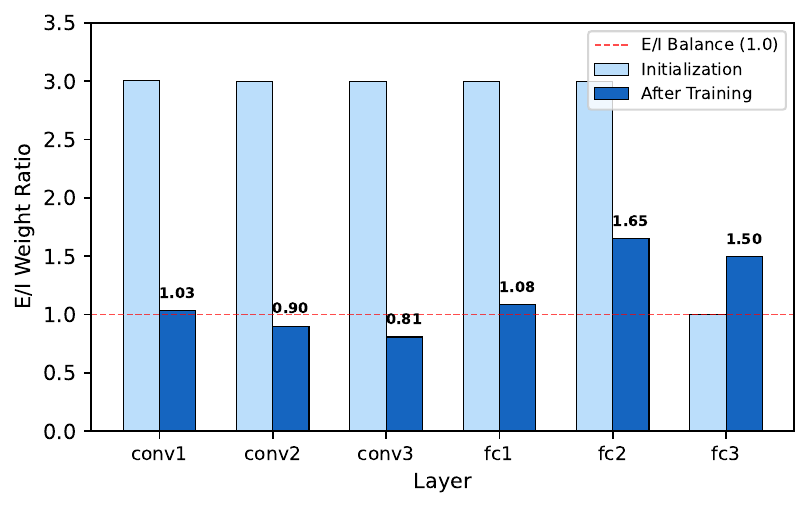}
        \caption{E/I weight ratio at init (3:1 for hidden layers) vs.\ after 250 epochs. Hidden layers converge toward balance; deeper layers become inhibitory-dominant.}
        \label{fig:ei_balance}
    \end{subfigure}
    \caption{Post-hoc analyses of the proposed ED model on CIFAR-10. (a)~Local surrogate gradient attenuation across layers. (b)~Emergent E/I balance from asymmetric initialization toward biological-like equilibrium.}
    \label{fig:posthoc}
\end{figure*}

\paragraph{Surrogate gradient attenuation.} Post-hoc analysis of training dynamics on CIFAR-10 (Figure~\ref{fig:posthoc}a) reveals local surrogate gradient magnitudes dropping from $6.6 \times 10^{-3}$ at the output layer to $2.7 \times 10^{-4}$ at the first hidden layer---a $25\times$ attenuation. This pattern is visible from epoch 3, indicating a structural property of sigmoid-gated error diffusion that motivates layer-specific widths. Per-layer learning rate multipliers ($3.0\times$ early, $0.5\times$ output) partially compensate but cannot substitute for wider sigmoid derivatives. The near-zero generalization gap ($0.98\%$) despite ${\sim}$32M nominal parameters is consistent with underfitting rather than overfitting, but the cause is ambiguous: the 37.3\% of weights at floor ($10^{-4}$) reduces effective capacity well below the nominal parameter count, so the limitation may reflect constrained capacity rather than (or in addition to) credit assignment quality.

\paragraph{Emergent E/I balance.} Weight-level E/I ratios (Figure~\ref{fig:posthoc}b) reveal that training drives the 3:1 asymmetric initialization toward near-balanced ratios (${\sim}1.0$) in hidden layers. A depth-dependent gradient emerges: the first layer reaches near-perfect balance (E/I = 1.03), the second becomes slightly inhibitory-dominant (0.90), and the third develops the strongest inhibitory bias (0.81). This increasing inhibition with depth is loosely consistent with biological observations that inhibitory circuitry and E/I balance vary systematically across cortical hierarchies \citep{markram2004interneurons}, though weight-level E/I ratios are an indirect proxy for biological E/I balance, which involves cell counts, firing rates, and synaptic strengths jointly. The convergence toward balance explains the task-dependent importance of asymmetric initialization: on MNIST, balance is reached quickly regardless of initial conditions ($\Delta = +0.0$ pp); on CIFAR-10, the asymmetric head start prevents early instability ($\Delta = -5.5$ pp).

\paragraph{Implicit sparsity.} The non-negative weight floor induces substantial implicit sparsity: 37.3\% of weights reach the floor ($10^{-4}$) after training. Inhibitory (cross-stream) FC connections are pruned most aggressively (up to 68.8\% at the floor), compared to 26--49\% for excitatory connections. Convolutional layers are much less affected ($<$1\% to 18\%). This asymmetric pruning suggests that the non-negative Dale-style parameterization introduces a structured capacity bottleneck that preferentially suppresses inhibitory connections in deeper layers.

\section{Discussion}

\paragraph{Summary.} The central finding of this work is that the dual-stream ED architecture is viable across both classification and reinforcement learning, but the auxiliary mechanisms required for competitive performance are domain-dependent. In classification, batch-centered class error and layer-specific sigmoid widths are critical stabilizers whose relative importance reverses between MNIST and CIFAR-10. In RL, the same dual-stream architecture with ReLU activations and RMS normalization, without sigmoid widths or batch-centering, achieves competitive performance on locomotion tasks, showing comparable but variable rewards on Ant, Humanoid, and HalfCheetah.

\paragraph{Implications for biological plausibility research.} Single-benchmark evaluations can produce misleading conclusions about which architectural features matter. The ablation reversal between MNIST and CIFAR-10 demonstrates that a component critical for one task can be negligible for another, and the DFA cross-domain pattern (strong on classification, weak on complex RL) reinforces this point.
The emergent E/I balance convergence, from 3:1 toward ${\sim}$1:1 with a depth-dependent inhibitory gradient, provides a connection to biological self-organization. Cortical circuits are known to develop balanced E/I ratios during maturation through intrinsic and synaptic homeostatic mechanisms \citep{markram2004interneurons, turrigiano2011too}, and our results suggest that a similar homeostatic process can emerge from gradient-driven learning under Dale's principle constraints, without explicit balance-enforcing mechanisms.
The success of ED-PPO on locomotion tasks despite using ReLU rather than sigmoid activations raises an interesting question about the relationship between activation function choice and credit assignment quality. In classification, the sigmoid derivative explicitly gates the error signal, making the width parameter critical. In RL, ReLU's unbounded positive derivative may provide a natural gradient pathway that avoids the attenuation problem without requiring layer-specific tuning. This activation-dependent difference in error flow may partially explain why the classification-specific innovations are unnecessary in the RL setting.

\vspace{-0.25cm}
\paragraph{Limitations.} The classification innovations (sigmoid widths, MCE, asymmetric init) are specific to the classification setting; we do not claim they transfer to RL. ED-PPO shows substantially higher variance than BP-PPO across all environments, suggesting that the coarse error routing introduces stochasticity that may limit reliability. The Craftax shortfall ($-4.0$ reward vs BP-PPO) indicates that ED's modular credit assignment may be insufficient for tasks requiring fine-grained temporal reasoning. The accuracy gap between ED and DFA on classification ($-0.9$ to $-7.4$ pp) quantifies the current cost of enforcing Dale's principle. On Craftax, DFA underperforms ED-PPO, suggesting that the cost of Dale's principle is task-dependent. Finally, the dual-stream architecture's ${\sim}4\times$ parameter overhead may itself limit effective capacity, particularly on classification where 37.3\% of weights are pruned to the floor.
From the perspective of adaptive computation, the emergent behaviors we observe---E/I balance self-organization, asymmetric pathway pruning, and task-dependent component criticality---suggest that biologically constrained architectures may develop internal regulatory mechanisms analogous to those found in living systems.

\paragraph{Future Work.} 
Several directions follow naturally from the results presented here. First, Dale's principle may offer hardware advantages by constraining synaptic weights themselves to be non-negative. In our Error Diffusion formulation, negative contributions are still required, but their sign is determined by fixed excitatory/inhibitory pathway structure rather than by arbitrary signed weights. Thus, the benefit is not the elimination of subtraction or differential circuitry altogether, but the replacement of unconstrained signed synapses with non-negative synaptic magnitudes and fixed-sign routing. This may be especially relevant for analog, photonic, or synapse-device-based neuromorphic substrates, where physical synaptic elements often naturally encode non-negative quantities, while sign can be implemented at the level of population identity, optical/electrical phase, or excitatory/inhibitory summation. This complements prior demonstrations of bioplausible rules on photonic hardware \citep{filipovich2022silicon}.
Second, the implicit sparsity we observe suggests that Dale-compliant training may yield model compression ``for free,'' since weights bounded below at zero cannot recover once suppressed, turning the floor into a natural pruning mechanism that could be exploited with structured sparsity kernels at inference time. Third, we conjecture that the dual-stream architecture may be unusually well-suited to continual and open-ended learning settings, where catastrophic forgetting is a central obstacle: the dedicated inhibitory stream provides a structural mechanism for dampening large gradient excursions, and the sign constraint prevents the kind of unconstrained weight sign flips that are thought to contribute to representational overwriting. Evaluating ED on sequential task streams and lifelong RL benchmarks is a natural next step. Fourth, the segregation of computation into excitatory ``amplifiers'' and inhibitory ``suppressors'' offers a potential interpretability handle that standard networks lack: when a model makes an error, one can in principle trace whether the mistake originated from a feature being falsely amplified by the excitatory stream or insufficiently suppressed by the inhibitory stream, providing a more mechanistic form of attribution than gradient-based saliency on sign-unconstrained networks. Finally, closing the accuracy gap to DFA and BP-PPO on the hardest tasks, whether through richer error-routing schemes, learned (rather than fixed modular) output-to-hidden projections, or hybrid activation functions that preserve sigmoid locality while mitigating attenuation, remains the most direct path toward making Dale-compliant learning competitive with unconstrained backpropagation.

\section{Conclusion}

We demonstrated that a dual-stream excitatory/inhibitory architecture trained with Error Diffusion achieves competitive performance across both supervised classification and reinforcement learning while maintaining non-negative weights consistent with Dale's principle. In the classification setting, three domain-specific innovations enable scaling from binary to 10-class problems, and their cross-task ablation reversal reveals that credit-assignment bottlenecks shift qualitatively with task difficulty. In reinforcement learning, the same core architecture integrated with PPO achieves comparable rewards to backpropagation-based PPO on locomotion tasks, without requiring the classification-specific stabilizers, though with substantially higher variance. The cross-domain pattern, competitive performance with domain-dependent auxiliary mechanisms, suggests that biologically plausible learning rules need not be monolithic: a shared architectural core can support diverse task demands when augmented with appropriate domain-specific components. The accuracy gap relative to BP-PPO on the hardest tasks quantifies the current cost of Dale's principle compliance and provides concrete benchmarks for future work on narrowing this gap while preserving biological fidelity.

\section{Acknowledgement}

Generative AI tools were used in the preparation of the manuscript, including text editing, code generation, image generation, and data analysis assistance. Additionally, we leveraged AI to generate ideas, run autonomous experimentation, and refine hypotheses. All claims, code implementation, manuscript writing, and figures were either reviewed by or created by the authors.

\footnotesize
\bibliographystyle{apalike}
\bibliography{references}

\end{document}

%% file: paper_authors.tex
Yutaro Yamada, Luca Grillotti, Rujikorn Charakorn, Sebastian Risi, David Ha, Robert Tjarko Lange\\
\mbox{}\\
Sakana AI\\
Corresponding authors: yutaro.yamada.y@gmail.com, robert@sakana.ai